\documentclass[letterpaper, 10 pt, conference]{ieeeconf}  

\IEEEoverridecommandlockouts                              

\makeatletter
\let\NAT@parse\undefined
\makeatother

\usepackage[sort&compress,numbers,sectionbib]{natbib}
\usepackage{balance}
\usepackage{graphicx} 
\usepackage{amsmath} 
\usepackage{amssymb}  
\usepackage{algorithm}
\usepackage[dvipsnames]{xcolor}
\usepackage{algpseudocode}
\usepackage{subfig}
\usepackage{multirow}
\usepackage{booktabs}
\usepackage{pifont}
\newcommand{\cmark}{\ding{51}}%
\newcommand{\xmark}{\ding{55}}%

\usepackage{soul}

\usepackage{hyperref}
\hypersetup{bookmarksopen,bookmarksnumbered,
pdfpagemode=UseOutlines,
colorlinks=true,
linkcolor=teal,
anchorcolor=teal,
citecolor=teal,
filecolor=teal,
menucolor=teal,
urlcolor=teal
}

\usepackage{balance}

\title{\LARGE \bf
Discovering Object Attributes by Prompting Large Language Models with Perception-Action APIs
}

\author{Angelos Mavrogiannis, Dehao Yuan, Yiannis Aloimonos
\thanks{
The authors are with the Department of Computer Science, University of Maryland, College Park, 8125 Paint Branch Dr, College Park, MD 20742, USA. {\tt\small {\{angelosm, dhyuan, jyaloimo\}}@umd.edu}}
\thanks{Our code, dataset, and robot demonstration video can be found here: \href{https://github.com/angmavrogiannis/Embodied-Attribute-Detection}{https://github.com/angmavrogiannis/Embodied-Attribute-Detection}.}
}

\begin{document}

\maketitle
\thispagestyle{empty}
\pagestyle{empty}

\begin{abstract}

There has been a lot of interest in grounding natural language to physical entities through visual context. While Vision Language Models (VLMs) can ground linguistic instructions to visual sensory information, they struggle with grounding non-visual attributes, like the weight of an object. Our key insight is that non-visual attribute detection can be effectively achieved by active perception guided by visual reasoning. To this end, we present a perception-action API that consists of VLMs and Large Language Models (LLMs) as backbones, together with a set of robot control functions. When prompted with this API and a natural language query, an LLM generates a program to actively identify attributes given an input image. Offline testing on the Odd-One-Out (O\textsuperscript{3}) dataset demonstrates that our framework outperforms vanilla VLMs in detecting attributes like relative object location, size, and weight. Online testing in realistic household scenes on AI2-THOR and a real robot demonstration on a DJI RoboMaster EP robot highlight the efficacy of our approach.

\end{abstract}

\section{INTRODUCTION}
Robots may need to understand natural language and execute verbal instructions from novice users to be valuable assistants in a household. Connecting natural language instructions to the physical world often requires robots to detect object attributes in order to discriminate between candidate objects. This is a challenging problem as instructions might be linguistically ambiguous, for example “Can you please get me the second mug from the right on that shelf?”. Identifying attributes can also be required implicitly to determine the state or affordance~\citep{gibson2014ecological} of an object in order to verify the feasibility of an action. These attributes might not be directly perceivable through vision sensors, for example “Is this lightweight enough to pick up?”.

In this paper, we focus on identifying object attributes in a programmatic fashion. Attributes usually appear in the form of descriptive adjectives, some of which might not be adequately represented in the training sets of data-driven perceptual models. Furthermore, while attributes might not necessarily characterize an object in an absolute scale, they are often applicable based on context~\citep{parikh2011relative}. We argue that attribute detection is highly contextual - an analogy of J. R. Firth's famous quotation ``You shall know a word by the company it keeps"~\citep{firth1957synopsis} applies to attributes. More specifically, characterizing an object as big, tall, or heavy can sometimes depend on the other objects and their respective attribute values in the current environmental context. Ambiguity can also arise due to occlusion or partial observability~\citep{zhang2023multimodal}. These problems might not occur when studying attribute detection in static images, but can be common in a household environment where a robot is tasked with executing user instructions. In these cases, erroneous attribute detection can be detrimental, producing and executing an action plan involving an incorrect object, or misinterpreting the affordance of an object and failing to even execute the action.

\begin{figure}
    \centering
    \includegraphics[width=\columnwidth]{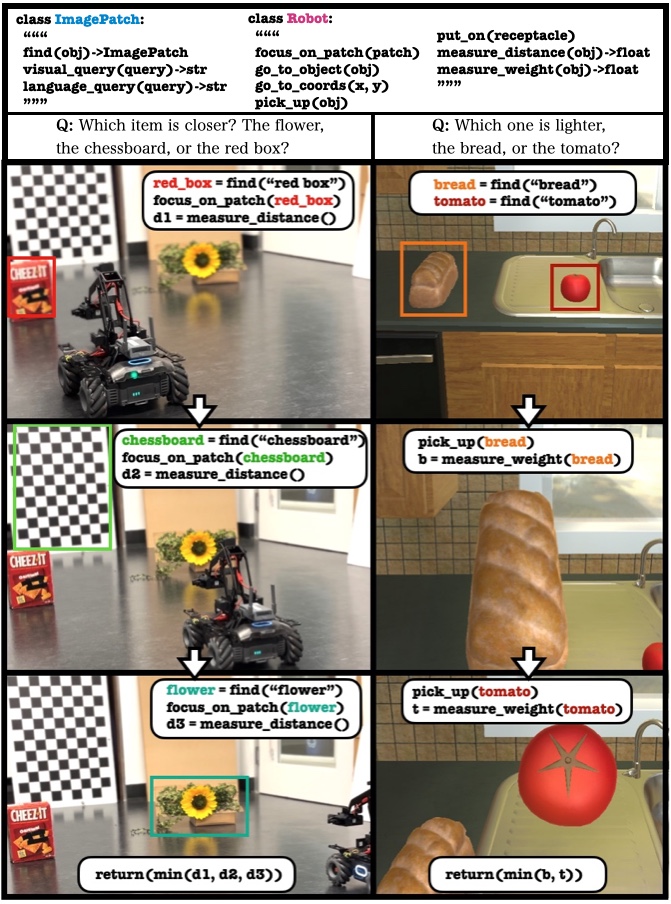}
    \caption{Demonstration of our perception-action API solving a minimum distance query on a real robot (\textit{left}) and a minimum weight query in simulation (\textit{right}). The LLM receives a perception-action API and a natural language query as input (\textit{top}). It then generates code that invokes API functions leveraging on-board sensors (camera, distance sensor, force/torque sensor) to actively identify these attributes.}
    \label{fig:distance}
\end{figure}

Existing attribute detectors \citep{pham2021learning,zhang2023multimodal,bravo2023open,chen2023ovarnet,guo2023lowa,tatiya2023mosaic,bianchi2023devil} are mainly obtained by either supervised training \cite{russakovsky2012attribute} or contrastive pre-training \cite{radford2021learning}. While attribute detection is an active area of research, it is often studied separately from embodied reasoning. 
To bridge this gap, we model attribute detection as visual reasoning with programs. This provides us with a powerful representation for reasoning in the presence of embodied agents and allows us to utilize the space of plans and movements via robot actions as programs~\citep{singh2023progprompt}.
Summarizing these ideas, our main observation is that modern real-world robotic systems relying on visually-driven attribute detection using VLMs in isolation can be myopic in language grounding. Our key insight is that combining different VLMs as visual reasoning functions with a robot control API can benefit from the code synthesis and commonsense reasoning capabilities of LLMs to actively reason about attribute detection in the form of computer programs. We prompt an LLM with an attribute detection API on a dataset that we curate, consisting of embodiment-crucial \textbf{location}-, \textbf{size}-, and \textbf{weight}-related attributes and construct a perception-action API for active attribute detection. Our key contributions can be summarized as follows:
\begin{itemize}
    \item We highlight some of the drawbacks of using VLMs for attribute detection in isolation and the complementary reasoning capabilities that emerge by reasoning in the form of LLM-generated visual programs.
    \item We construct a perception-action API by integrating visual reasoning with robot control functions and demonstrate its benefits by invoking active perception behaviors towards solving attribute detection queries.
    \item We release an end-to-end framework that integrates this perception-action API on a real robotic platform using visual servoing.
\end{itemize}
\section{RELATED WORK}

\textbf{Attribute Detection:} Attribute detection has been a fundamental problem in the computer vision community with early work~\citep{NIPS2007_ed265bc9,farhadi2009describing,lampert2009learning,russakovsky2012attribute} on learning visual attribute classifiers to describe unseen objects. There has been work on identifying relative attributes~\citep{parikh2011relative,chen2018compare} but these approaches require prior training and can be limited to a certain domain (\citep{chen2018compare} demonstrate relative attributes of shoes and facial characteristics), while we show a general method that works in a zero-shot fashion. Recent approaches have applied Open-Vocabulary object Detection (OVD)~\citep{zareian2021open} to attribute detection~\citep{bravo2023open,chen2023ovarnet}. The goal in OVD is to detect unseen classes of objects defined at inference time in the form of textual queries. Bravo et al.~\cite{bravo2023open} showed that the performance of various VLMs in zero-shot attribute detection is still low compared to OVD. However, most of these approaches focus on visually-perceivable attributes in disembodied settings. On the other hand, we focus on embodiment-crucial attributes such as the weight of an object, leveraging the physical reasoning capabilities of VLMs~\citep{wang-etal-2023-newton}. Our simulations in AI2-THOR and our robot demonstration shows that our end-to-end framework can invoke active perception behaviors to reason about object attributes, inspired by prior work~\citep{zhang2023multimodal,tatiya2023mosaic}.
\begin{figure*}[t]
    \centering
    \includegraphics[width=\textwidth]{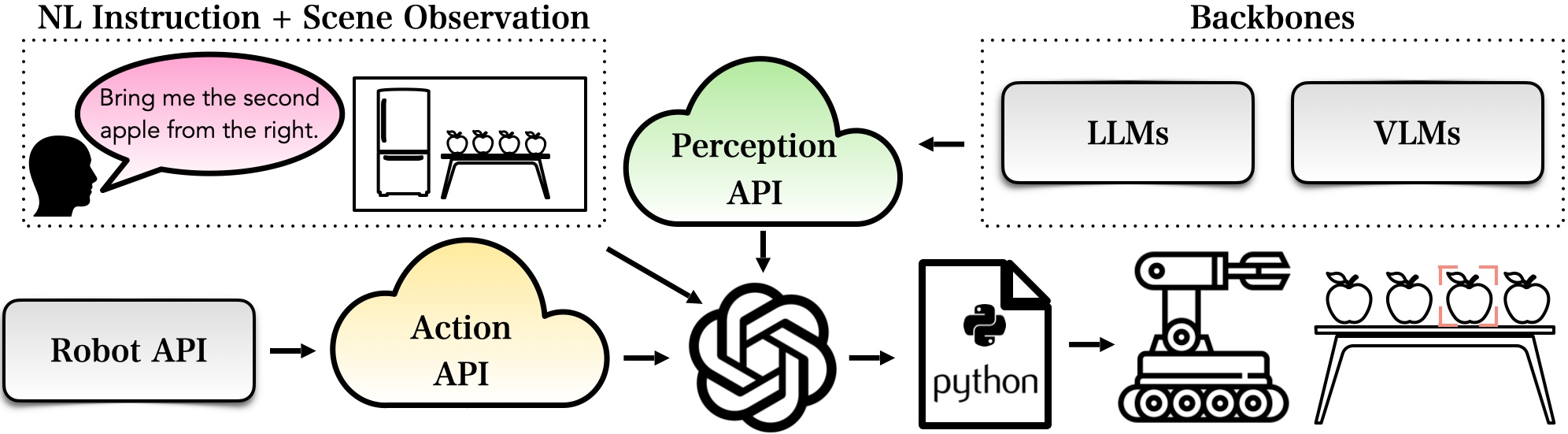}
    \caption{We describe our end-to-end framework for embodied attribute detection. The LLM receives as input a perception API with LLMs and VLMs as backbones, an action API based on a Robot Control API, a natural language (NL) instruction from a user, and a visual scene observation. It then produces a python program that combines LLM and VLM function calls with robot actions to actively reason about attribute detection.}
    \label{fig:pipeline}
\end{figure*}

\textbf{LLMs as Embodied Agents:} LLMs have shown remarkable performance in translating natural language instructions to robot actions that are admissible in a given environment~\citep{huang2022language,pmlr-v205-ichter23a,huang2022inner,vemprala2023chatgpt,mavrogiannis2023cook2ltl,liang2023code,huang2023instruct2act}. This has been achieved predominantly by a few-shot prompting scheme where the LLM receives a set of examples of sample tasks and action plans as input, and generates an action plan for an unseen task at inference time. While some of these works rely on the inherent commonsense reasoning capabilities of LLMs to map natural language to actions~\citep{huang2022language,huang2022inner,pmlr-v205-ichter23a}, a line of work expressing action plans as programs~\citep{singh2023progprompt,vemprala2023chatgpt,liang2023code,huang2023instruct2act} has been using programming language constructs to elicit more profound reasoning capabilities such as action precondition checking through conditional and assertion statements~\citep{singh2023progprompt}, reasoning about task execution using control flow tools~\citep{vemprala2023chatgpt} and recursively defining undefined functions~\citep{liang2023code}, or simply invoking VLMs in task execution functions~\citep{huang2023instruct2act}. However, these approaches focus on action planning using OVD methods in isolation~\citep{gu2021open,kamath2021mdetr,kirillov2023segment,radford2021learning} and might not be able to handle attribute detection in challenging scenarios. Similar to~\citep{biggie2023tell}, our approach combines the expressiveness of an intermediate programmatic representation and the complementary reasoning capabilites of LLMs and VLMs~\citep{zeng2022socratic} to reason about attribute detection under an LLM-prompting scheme.

\textbf{Visual Reasoning with Programs:} Generating and executing programs for vision applications originated from Neural Module Networks (NMNs)~\citep{andreas2016neural,hu2017learning,johnson2017inferring}, on the basis of the idea that complex vision tasks are fundamentally compositional. Motivated by this idea, NMNs decompose a task into trainable modules that learn specific perceptual functions. However, these models produce domain-limited programs, rely on hand-tuned parsers~\citep{andreas2016neural} or are difficult to optimize~\citep{hu2017learning,johnson2017inferring}. To overcome these shortcomings, a recent line of work has proposed a formulation of generating visual programs to deal with image-based natural language queries through in-context learning with an LLM. The programs consist of pseudocode instructions~\citep{gupta2023visual} or executable python code~\cite{suris2023vipergpt,subramanian2023modular,stanic2024towards} and intermediate variables that map to computer vision models, image processing subroutines, or LLMs. These intermediate variables are consumable downstream and illustrate a step-by-step reasoning process towards the task at hand, which is primarily related to language grounding or Visual Question Answering (VQA). Our approach invokes active perception robot behaviors guided by visual programming towards attribute detection.

\section{METHOD}
\subsection{Problem Statement}
Consider a robot equipped with a set of sensors $\mathcal{S}$ in a scene with a set of objects $\mathcal{O}$. The robot is tasked with executing a natural language instruction $inst=f(a,g,o, img)$, where $a$ is a high-level action, $g$ is an object attribute, $o$ is an object, and $img$ is an input image. Our goal is to determine whether an object exists with this attribute, expressed by the predicate $g(o)$, and localize it in $img$ by obtaining its bounding box coordinates $\mathcal{X}=\{x_{min},y_{min},x_{max},y_{max}\}$. If $\exists o$, such that $g(o)$ holds, a visual navigation policy $\pi(a(\mathcal{X}))$ is deployed, which allows the robot to leverage its sensors $S$, navigate and manipulate object $o$ given its 2D bounding box coordinates $\mathcal{X}$ towards task completion.

We adopt a generalized definition of an attribute, viewing it as an abstract property of an object that does not necessarily map to a visual representation, including primarily descriptive adjectives related to the size (\textit{big}) or weight (\textit{heavy}) of an object, but also prepositional phrases indicating spatial relationships (\textit{the second object from the left}).

\subsection{Prompt-based Attribute Detection}
We adopt the methodology of Surís et al.~\cite{suris2023vipergpt} into constructing a Python API for attribute detection. The API consists of a main \texttt{ImagePatch} class that is instantiated by an input image $img$. \texttt{find} is the fundamental function of the API that uses an OVD model (GLIP~\citep{li2022grounded}) to locate an object $o$ and return the detection-resulting cropped patch $\mathcal{X}$ from the image given a chunk of natural language $inst$. While we do not explicitly define a function for spatial reasoning, we provide in-context examples encoded in the docstring of the function. The examples perform pixelwise math given bounding box coordinates $\mathcal{X}$ of detected objects $\mathcal{O}$ that are returned from calls to \texttt{find} in order to reason about relative object locations on the image frame. \texttt{visual\_query} calls a pre-trained VLM (BLIP-2~\citep{li2023blip}) to provide a textual answer to a visual query given an image. \texttt{language\_query} recursively calls the LLM with a textual query such as the visually-extracted return value from \texttt{visual\_query}. This API can be viewed as an internal dialogue between VLMs and LLMs, similar to the idea of Socratic Models~\citep{zeng2022socratic}, but enhanced by structural programming tools hosted on a pythonic platform. In the following subsections we describe the complementary reasoning capabilities that emerge from calling these functions.

\subsection{Programmatic Reasoning}
Reasoning in the form of programs inherits the expressiveness of programming languages through control flow tools, data structures, and built-in methods. LLM-generated programs invoke loops to iterate over detected object patches and conditional statements to determine whether an object exists ($\exists o$) in the input image $img$ and whether it possesses an attribute $g$, grounding the predicate $g(o)$. Python lists are used to store instances of image patches dynamically with the \texttt{append} function. Other built-in functions such as \texttt{sort} and the \texttt{lambda} function utilize horizontal and vertical coordinates of the detected bounding boxes and their centroids within simple mathematical operations and leverage basic geometrical notions (e.g. computing the area of an image patch) to reason about the size or relative position of detected entities. The generated code is interpretable and mainly consists of elementary arithmetic in the image frame. The commonsense reasoning component of the LLM proves to be essential in mapping complex language queries to these computations, as well as adapting attribute interpretation to image-specific contexts, as shown in Sec.~\ref{sec:results}.

\subsection{Vision-Informed Language Reasoning}
Combining LLMs and VLMs in the input API prompt unlocks complementary reasoning capabilities through information passing between different model calls in the the context of an LLM-generated program. This model interplay can be particularly efficient when dealing with non-visually perceivable attributes, such as estimating the weight of an object. In this case, \texttt{visual\_query} can serve as a zero-shot object recognition function, subsequently passing information to the input of the \texttt{language\_query}, which deduces factual knowledge on the weight of the recognized object, as explained in Sec~\ref{non-visual}. Similarly, when a task instruction involves an attribute that does not typically describe an object in an absolute scale, recognizing adjacent objects in addition to the object at hand establishes context. Then, posing a textual query with this visually-acquired information might reduce ambiguity and provide the correct grounding, leveraging the domain knowledge of LLMs. We demonstrate an empirical evaluation of such use cases in Sec.~\ref{sec:evaluation}.

\subsection{Embodied Attribute Detection}
Attribute detection in embodied settings often requires active perception. To this end, we formulate an action-perception API by integrating the attribute detection API with a high-level robot control API (See Fig.~\ref{fig:pipeline}). The robot control API is implemented as a \texttt{Robot} Python class that consists of sensors as member variables and methods that map to simple navigation and pick-and-place actions. The robot can navigate to an object with the \texttt{go\_to\_object} function which implements a visual navigation policy by calling \texttt{go\_to\_coords} with an image patch $\mathcal{X}$ as a parameter. \texttt{pick\_up} and \texttt{put\_on} implement picking and placing actions. Assuming the lack of an on-board RGB-D camera or a depth estimation model, a robot could employ additional sensors to measure the distance to an object in order to reason about scene geometry or depth. This can be achieved by the \texttt{measure\_distance} function which calls \texttt{focus\_on\_patch}, a function that aligns the geometric center of the image frame to an object patch and can then retrieve the distance sensor measurement to compute the distance from the camera to that object. A demonstration on a real robot is shown in Fig.~\ref{fig:distance} (\textit{left}). Similarly, \texttt{measure\_weight} can measure the weight of an object grasped by the robot, under the precondition that the robot first navigates and picks it up. These preconditions are encoded in the example use of the function in a docstring. We integrate this perception-action API into an AI2-THOR simulated environment and a real robot and demonstrate its benefits in Sec.~\ref{sec:results}.

\section{EVALUATION}
\label{sec:evaluation}
We design a set of experiments to showcase some of the drawbacks of using OVD or VQA models for attribute grounding in isolation and we highlight the complementary commonsense reasoning that emerges by visual reasoning with LLM-generated programs including actions.

\subsection{Spatial Reasoning}

We evaluate the spatial reasoning capabilities of the attribute detection API by comparing its ability to ground linguistically complex spatial queries with an open-vocabulary object detector~\citep{li2022grounded}. 
We manually craft a dataset that consists of $200$ challenging spatial queries based on the Odd-One-Out (O\textsuperscript{3}) Dataset~\citep{Kotseruba2019BMVC}. Every image in this dataset includes multiple instances of an object or similar objects with an instance being slightly different to stand out. We leverage the multiple instances of an object to invoke reasoning that requires differentiating between objects based on their relative attributes rather than obvious qualitative differences between entirely different objects. Therefore, instead of focusing on relative attributes that localize the object with respect to another object of different type in the image~\citep{kazemzadeh2014referitgame,suris2023vipergpt} (e.g. ``\textit{the car to the left of the tree}”), our $100$ \textbf{location} queries require commonsense reasoning in the form of counting and establishing the relative order of an arranged set of objects, such as ``\textit{second umbrella from the left at the second to last row}" or ``\textit{the window in the middle at the bottom}". Our $100$ \textbf{size} queries utilize descriptive size-related adjectives (long, wide, short, large etc.) in their superlative and absolute form, such as ``\textit{the tallest item}" or ``\textit{the wide line}", respectively. We test the same queries on both forms and expect that the superlative form will outperform the absolute, forcing a specific object to stand out by emphasizing its attribute. We anticipate that the attribute detection API will outperform OVD by incorporating pixelwise mathematical operations and expressive python utility functions.

\subsection{Non-visually Perceivable Attributes}
\label{non-visual}
\begin{figure}
    \centering
    \includegraphics[width=\columnwidth]{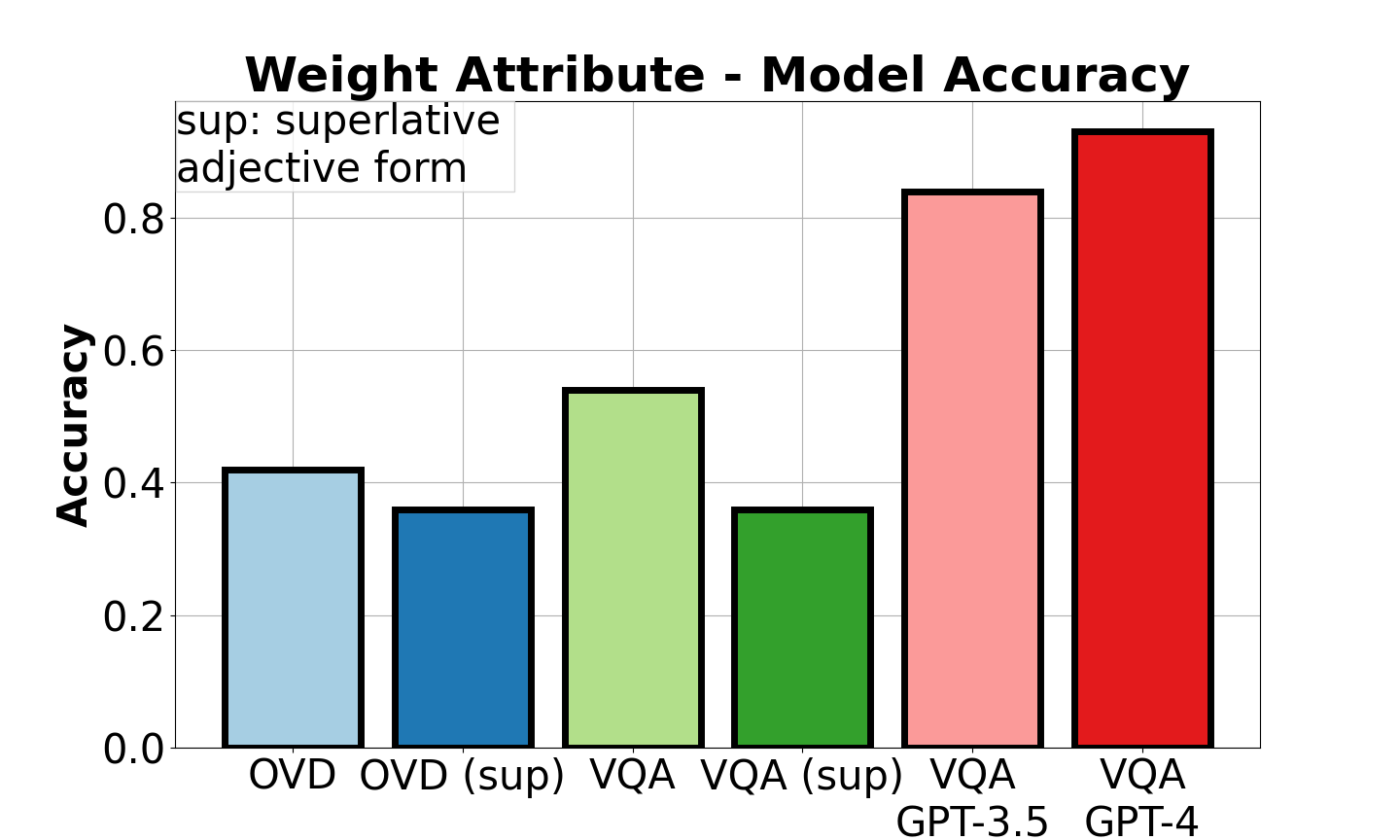}
    \caption{The accuracy of OVD (GLIP), VQA (BLIP-2), and VQA$+$GPT in determining the heaviest object in an image.}
    \label{fig:weight_bar}
\end{figure}
\begin{figure}
    \centering
    \includegraphics[width=\columnwidth]{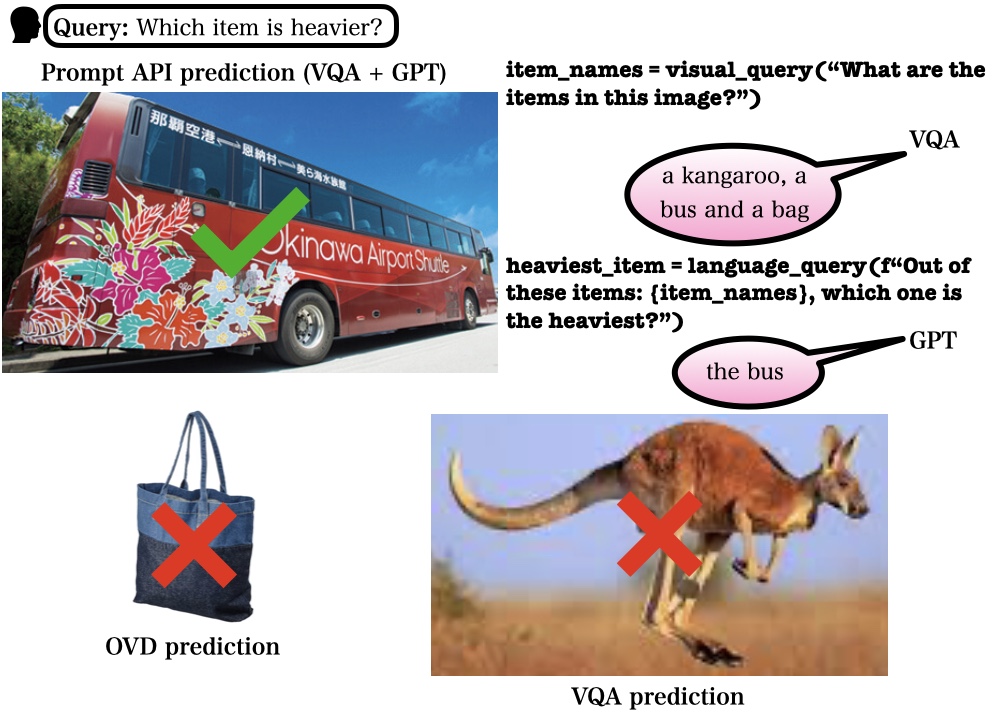}
    \caption{An example where OVD and VQA fail to identify the heaviest object in the image (\textcolor{Red}{\xmark}), but the API prompt-generated code (VQA$+$GPT) returns the correct answer (\textcolor{Green}{\cmark}).}
    \label{fig:weight_qualitative}
\end{figure}

To evaluate more profound reasoning capabilities, we focus on the \textbf{weight} of an object as a representative sample of non-visually perceivable attributes, since it is crucial for executing essential manipulation tasks. In the absence of a dataset with a suitable schema for our use case, we prompt GPT-4 to generate sets of objects of different weight, along with the ground truth label of the heaviest object. To simplify the task, we design the prompt so that the object weight distribution is monotonic and clearly distinguishable by a human observer: ``\textit{Generate $100$ triplets of objects where each object is significantly heavier than the other (for example: feather, dog, car)}." After acquiring the generated textual data, we utilize it to extract relevant images from the web and arrange them to form an image dataset where each data sample is an image that includes three objects of monotonically decreasing weight. Assuming some rudimentary commonsense reasoning functionality in VLMs~\citep{brohan2023rt}, we expect that they are capable of identifying obvious differences in weight and hence dealing with examples that are intuitive to humans, such as selecting the heaviest object between a handbag, a kangaroo, and a bus (See Fig.~\ref{fig:weight_qualitative}). We compare the performance of OVD: \texttt{find(}\textit{``a heavy object"}\texttt{)}, VQA: \texttt{visual\_query(}\textit{``Out of these items, which one is the heaviest?"}\texttt{)}, and vision-informed language reasoning (VQA$+$GPT) that is invoked by our prompt API: \texttt{visual\_query(}\textit{``What are the items in this image?"}\texttt{)} $\rightarrow$ \texttt{language\_query(}\textit{``Out of these items, which one is more likely to be the heaviest one?"}\texttt{)}.

\subsection{Evaluation in Embodied Settings}
To evaluate our perception-action API in embodied settings, we adapt it to a simulated AI2-THOR~\citep{kolve2017ai2} household environment. We assume that the robot comes with a proximity sensor and a force/torque sensor mounted on the wrist of the gripper, capable of measuring the weight of an object. To replicate the behavior of these sensors, we query the simulator for the distance between an object centered on the frame captured by the on-board robot camera, and build a queryable dictionary that maps an object to an approximate weight when the robot is holding that object. We measure the accuracy (\%) of our perception-action API in estimating the relative distances of objects from the robot camera and identifying the most lightweight object. Our baselines are OVD, VQA, GPT-4o, and the attribute detection API (VQA/OVD+GPT-4). We use the following prompt templates: ``\textit{Out of the \{objects\}, which one is closer to me?}", ``\textit{Out of the \{objects\}, which one is the most lightweight?}". We anticipate that the LLM-generated programs from the perception-action API are capable of actively interacting with the environment to identify object attributes leveraging sensor-powered visual reasoning functions and robot actions.

\subsection{Hypotheses}
We formalize these insights into the following hypotheses:

\textbf{H1:} OVD$+$GPT outperforms OVD- and VQA-only baselines in \textbf{location}- and \textbf{size}-related queries.

\textbf{H2:} VLMs possess the rudimentary reasoning capability to tackle evident \textbf{weight} estimation queries.

\textbf{H3:} The superlative form of a descriptive adjective yields a better grounding performance than the absolute form.

\textbf{H4:} Our perception-action API solves attribute detection queries by actively interacting with the environment.

To evaluate these hypotheses, we measure the grounding accuracy by comparing the bounding boxes returned by OVD and OVD+GPT. For OVD, we report results from GLIP~\citep{li2022grounded}. In the case of the weight attribute, we additionally consider the textual output of VQA$+$GPT. We demonstrate results from GPT-3.5, GPT-4, and GPT-4o in the embodied settings.



\subsection{Results \& Discussion}
\label{sec:results}
\begin{figure}
    \centering
    \begin{minipage}{0.5\linewidth}
        \centering
        \includegraphics[width=\linewidth]{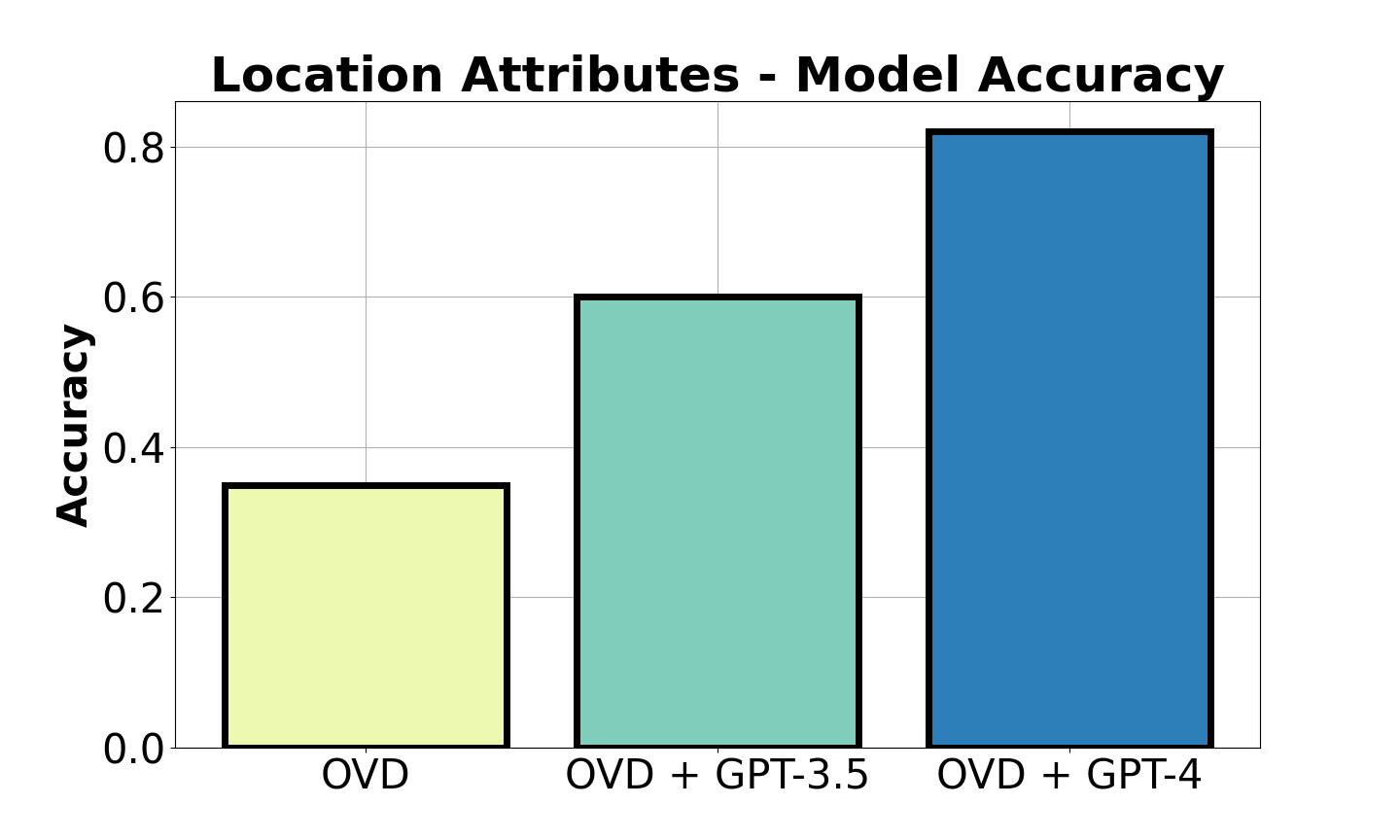}
    \end{minipage}\hfill
    \begin{minipage}{0.5\linewidth}
        \centering
        \includegraphics[width=\linewidth]{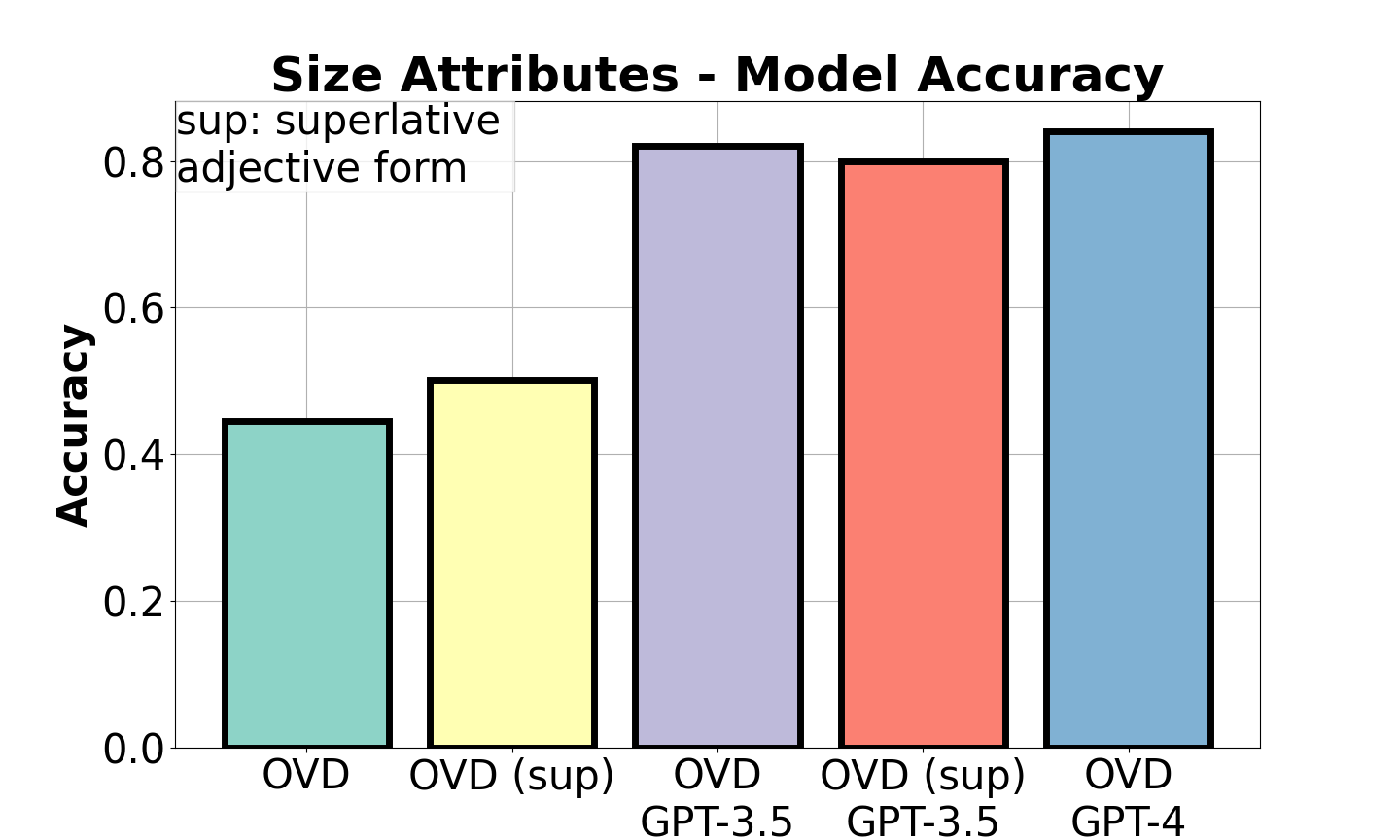}
    \end{minipage}
    \caption{We compare the accuracy of OVD-only (GLIP) with (OVD$+$GPT) on our \textbf{location} (\textit{left}) and \textbf{size} (\textit{right}) datasets.}
    \label{fig:location_size}
\end{figure}

Based on the results in Fig.~\ref{fig:weight_bar}, Fig.~\ref{fig:location_size}, and Table~\ref{tab: simulation_results} we make the following observations regarding our hypotheses.

\begin{figure}
    \centering
    \includegraphics[width=\columnwidth]{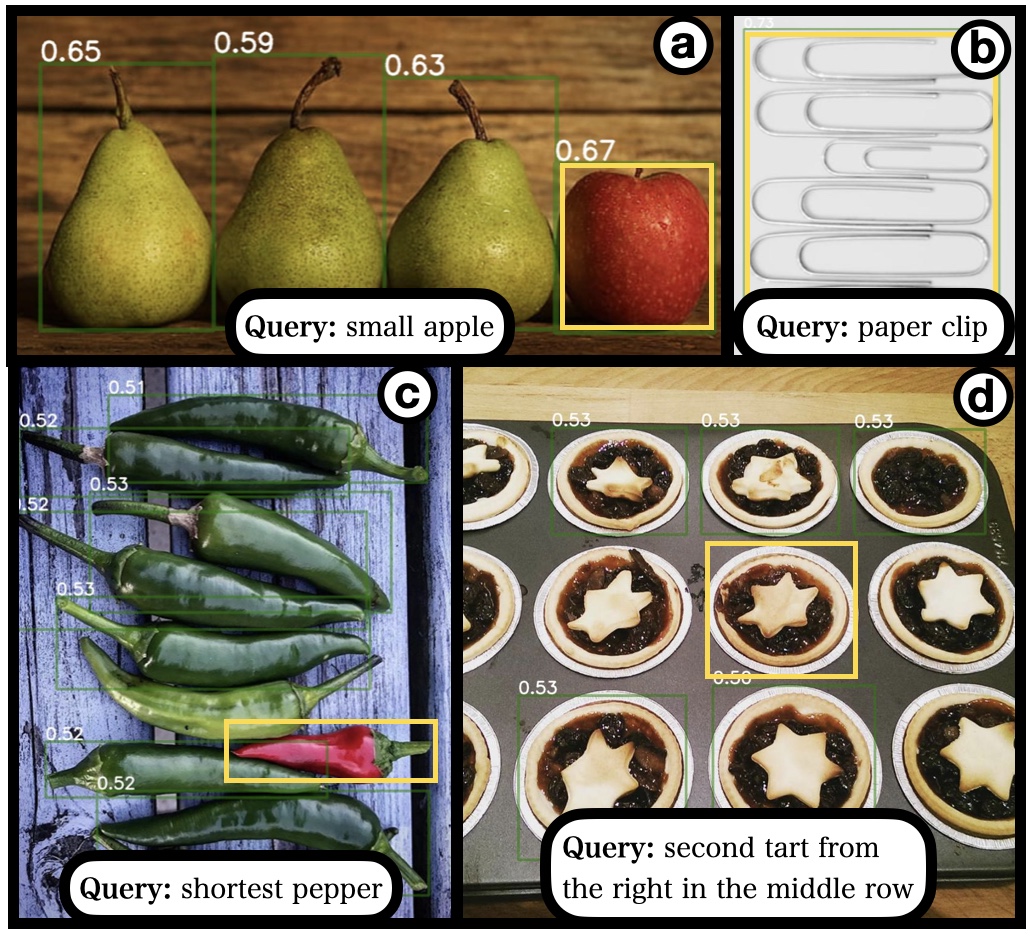}
    \caption{OVD (numbers in \colorbox{lightgray}{\textcolor{white}{white}} denote the OVD confidence score) and OVD$+$GPT predictions are shown with \textcolor{Green}{green} and \textcolor{Yellow}{yellow} bounding boxes, respectively. \textit{a} shows an example of agreement between OVD and OVD$+$GPT, \textit{b} a mutual failure case, and \textit{c}, \textit{d} show cases where OVD$+$GPT exhibits superior performance compared to OVD.}.
    \label{fig:loc_size_qualitative}
\end{figure}

\textbf{H1:} Our first hypothesis is confirmed. We find that OVD$+$GPT significantly outperforms OVD (See Fig.~\ref{fig:location_size}) in \textbf{location} queries by $134$\% and in \textbf{size} queries by $67$\%. Qualitative examples are shown in Fig.~\ref{fig:loc_size_qualitative}. In example \textit{a}, OVD and OVD$+$GPT correctly identify the apple as the small fruit. Example \textit{b} shows a common mutual failure case where the bounding box for one paper clip mistakenly includes all of them. We also observe failure cases due to occlusion. Examples \textit{c}, \textit{d} show cases where OVD$+$GPT outperforms OVD. In these examples, all instances of the object in the query are localized with the \texttt{find} function, and then the coordinates of the bounding boxes and their centroids are used to compute relative distances and areas and sort them, if needed. In terms of the \textbf{size}-related queries, GPT-4 can adapt the generated code to the context of an image and interpret what dimension \textit{long} or \textit{short} corresponds to based on the orientation of an object in an image. On the other hand, GPT-3.5 is more rigid and tends to tie certain adjectives to hardcoded dimensions following common norms. For example, in Fig.~\ref{fig:loc_size_qualitative} (\textit{c}), it cannot connect the adjective \textit{short} to the red pepper, since it is horizontally aligned. Such minor details explain the slight discrepancy in performance, which is still superior than vanilla OVD when identifying attributes with an attribute detection API. We would need to add targeted examples in the prompt API covering all failure cases to induce equal accuracy from both models. Finally, in example \textit{d}, OVD$+$GPT-4 understands that the arrangement of the tarts is forming a row and column pattern. On the contrary, OVD fails to recognize this pattern, and OVD$+$GPT-3.5 exhibits a context-agnostic interpretation of rows, dividing the image into parts based on the image height, which yields incorrect results.

\textbf{H2:} Our second hypothesis is only partially confirmed. Based on Fig.~\ref{fig:weight_bar}, the combination of VQA, followed by a call to an LLM, significantly outperforms OVD-only (+$121$\%) and VQA-only (+$72$\%) solutions although we expected that all the models would be able to handle simple comparative attribute detection tasks. We believe that the step-by-step reasoning process followed by the prompt API leverages the strength of each model separately. On the contrary, burdening a model with additional reasoning tasks upon the ones that it was naturally tasked with in the first place (zero-shot language-conditioned object detection for GLIP, and zero-shot object recognition for BLIP-2) might be the reason we are missing out on its full task-specific potential. Another potential reason for this discrepancy in performance is that the pre-training objective of these models might not be aligned to our specific use case, which is identifying non-visually perceivable attributes.

\textbf{H3:} Our third hypothesis is rejected, mainly because the absolute form of an adjective yields better performance in the weight estimation task (See Fig.~\ref{fig:weight_bar}). In Fig.~\ref{fig:weight_bar}, Fig.~\ref{fig:location_size}, \textit{sup} stands for superlative adjective form in the prompt. GPT-4 demonstrates the same performance for both forms, therefore we report a joint measurement for this model in Fig.~\ref{fig:location_size}.

\begin{table}
    \centering
    \resizebox{.7\columnwidth}{!}{%
    \begin{tabular}{l|l|l}
        \toprule
        \textbf{Method}& \multicolumn{2}{c}{\textbf{Task}}\\
        \hline
         & Weight& Distance\\
        \hline
        OVD & $0.14$ & $0.64$\\
        VQA & $0.64$ & $0.56$\\
        Attribute Detection API & $0.90$ & $0.22$\\
        GPT-4o & $0.88$ & $0.70$\\
        Perception-Action API & $\mathbf{0.96}$ & $\mathbf{0.94}$\\
        \bottomrule
    \end{tabular}
    }
\caption{Performance of our perception-action API in $50$ weight and $50$ distance estimation queries in simulated AI2-THOR~\citep{kolve2017ai2} household environments against baselines.}
\label{tab: simulation_results}
\end{table}

\textbf{H4:} Confirming our hypothesis, our perception-action API solves both tasks and outperforms all baselines based on the results shown in Table~\ref{tab: simulation_results}. In distance estimation, the robot first identifies an object patch with \texttt{find}, and then leverages its distance sensor by focusing on the detected image patch with \texttt{focus\_on\_patch} and calling \texttt{measure\_distance} to get the measurement. In weight estimation, after locating the patch with \texttt{find}, the robot navigates (\texttt{go\_to\_object}) and proceeds to \texttt{pick\_up} every object and measure its weight using the force/torque sensor by calling \texttt{measure\_weight}. At every measurement, the generated programs compare the currently measured value with a previously stored minimum and update it if the current value is lower, finally yielding the minimum distance or weight. In some cases, \texttt{find} cannot locate the object patches leading to failures. We believe that this occurs because of the image quality of the simulated objects, which are not as realistic as the ones in real-world images and the OVD model that \texttt{find} uses struggles to locate them. In distance estimation, the generated code by the attribute detection API (OVD+GPT-4) incorrectly hardcodes the distance from the object to the robot camera to the distance of the object from the geometric center of the image frame, leading to a very low accuracy.

\subsection{End-to-End Framework - Robot Demonstration}
We integrate our perception-action API into a real robot by implementing a wrapper over a robot-specific API. We deploy the combined end-to-end framework on DJI\textsuperscript{®} RoboMaster\textsuperscript{TM} EP~\citep{dji}, an affordable ground robot with holonomic movement and pick-and-place capabilities. A demonstration of our framework in action is shown in Fig.~\ref{fig:distance}.
Picking-and-placing an object first requires navigating in front of it with the appropriate orientation (\texttt{go\_to\_object}). To this end, we leverage sensory information to design a control policy for implementing the \texttt{go\_to\_object} function to navigate to a detected object. The control policy is further divided into two sub-policies: i) a visual servoing-based control policy for the lateral movement that aligns the center of the patch of the detected object to the center of the image frame captured by the on-board robot camera (\texttt{focus\_on\_patch}), ii) a control policy for the longitudinal movement that steers the robot towards a proximal position to the object at hand based on an infrared distance sensor. The target distance from the gripper to an object is a pre-computed functional gap, or in other words an experimentally determined \textit{sweet spot} for picking and placing. Each of these policies is separately handled by a hand-tuned Proportional-Integral-Derivative (PID) controller. The robot is connected to a (\textit{local}) computer via wifi connection and communicates with a (\textit{remote}) computing cluster through a client-server architecture running on an SSH tunnel. To reduce latency due to the computational load of deploying a VLM on the cluster, we only run OVD on the first frame captured by the robot camera, and then track the corresponding position(s) with the Kanade–Lucas–Tomasi (KLT) feature tracker\footnote{We follow the implementation in \href{https://github.com/ZheyuanXie/KLT-Feature-Tracking.git}{https://github.com/ZheyuanXie/KLT-Feature-Tracking.git}.}~\citep{tomasi1991detection}. 

\section{LIMITATIONS \& FUTURE WORK}
\textbf{Sensor Integration:} Our experiments provide some insights on failure cases and emerging reasoning capabilities of VLMs in attribute detection. We demonstrate the applicability of our action-perception API on a robot in simulation and in the real world. In the future we plan to leverage the compositionality of our API and extend its sensing capabilities by incorporating more sensors (e.g. IMU, temperature sensor) via wrapper functions, supporting the discovery of additional attributes through active perception.

\textbf{Error Propagation across Model Calls:} In Sec.~\ref{sec:results} we showed how the attribute API (VQA+GPT) outperforms calling an OVD or VQA model in isolation. However, if the first call yields an incorrect result, any downstream calls consume erroneous parameters and hence lead to an incorrect final result. In the future we plan to develop mechanisms that leverage additional feedback from the environment to catch such exceptions before errors propagate downstream.






\balance
\bibliographystyle{ieeetr}
\bibliography{refs}

\end{document}